\documentclass[preprint]{clear2025} 

\title[Do Large Language Models Show Biases in Causal Learning?]{Do Large Language Models Show Biases in Causal Learning?}
\usepackage{times}
\usepackage{float}
\usepackage{array}

\clearauthor{\Name{asdf} \Email{an1@sample.com}\\
 \Name{asdf} \Email{an2@sample.com}\\
 \Name{asdfadf3} \Email{an3@sample.com}\\
 \addr UBA}

\clearauthor{%
 \Name{María Victoria Carro} \Email{6381013@studenti.unige.it}\\
 \addr Università degli Studi di Genova, GE, Italy \&
FAIR, IALAB, University of Buenos Aires, BA, Argentina
 \AND
 \Name{Francisca Gauna Selasco} \Email{fgaunas@fi.uba.ar}\\
 \Name{Denise Alejandra Mester} \Email{mester330@est.derecho.uba.ar}\\
 \Name{Margarita Gonzáles} \Email{Gonzalezpereiro192@est.derecho.uba.ar}\\
 \addr FAIR, IALAB, University of Buenos Aires, BA, Argentina%
\AND
\Name{Mario A. Leiva} \Email{mario.leiva@cs.uns.edu.ar}\\
\addr Dept. of Computer Science and Engineering,
Universidad Nacional del Sur (UNS) \&
Inst. of Computer Science and Engineering
(ICIC UNS-CONICET), Bahía Blanca, BA, Argentina
\AND
\Name{Maria Vanina Martinez} \Email{vmartinez@iiia.csic.es}\\
\addr Artificial Intelligence Research Institute (IIIA-CSIC), Universidad Autónoma de Barcelona, Barcelona, España
\AND
\Name{Gerardo I. Simari} \Email{gis@cs.uns.edu.ar}\\
\addr Dept. of Computer Science and Engineering,
Universidad Nacional del Sur (UNS) \&
Inst. of Computer Science and Engineering
(ICIC UNS-CONICET), Bahía Blanca, BA, Argentina \&
School of Computing and Augmented Intelligence, {Arizona State University}, {USA}
}

\begin{document}

\maketitle

\begin{abstract}%
Causal learning is the cognitive process of developing the capability of making causal inferences based on available information, often guided by normative principles. 
This process is prone to errors and biases, such as the illusion of causality, in which people perceive a causal relationship between two variables despite lacking supporting evidence. This cognitive bias has been proposed to underlie many societal problems, including social prejudice, stereotype formation, misinformation, and superstitious thinking. In this research, we investigate whether large language models (LLMs) develop causal illusions, both in real-world and controlled laboratory contexts of causal learning and inference. To this end, we built a dataset of over 2K samples including purely correlational cases, situations with null contingency, and cases where temporal information excludes the possibility of causality by placing the potential effect before the cause. We then prompted the models to make statements or answer causal questions to evaluate their tendencies to infer causation erroneously in these structured settings. Our findings show a strong presence of causal illusion bias in LLMs. Specifically, in open-ended generation tasks involving spurious correlations, the models displayed bias at levels comparable to, or even lower than, those observed in similar studies on human subjects. However, when faced with null-contingency scenarios or temporal cues that negate causal relationships, where it was required to respond on a 0--100 scale, the models exhibited significantly higher bias. These findings suggest that the models have not uniformly, consistently, or reliably internalized the normative principles essential for accurate causal learning. %
\end{abstract}


\section{Introduction}

Causal learning is the ability to extract causal knowledge from available information \citep{Blanco2017}, perceiving two variables as causally related (e.g. smoking produces lung cancer). This cognitive process is informed by several features of the accessible evidence. In fact, there are some principles that should guide normative causal inference, including temporal ordering and contingency\footnote{Contingency refers to a conditional relationship where one event changes the probability of another, serving as a key cue in causal learning. Quantified by the metric delta P ($\Delta P$), similar to a correlation coefficient, it captures both the direction (generative or preventative) and strength of the relationship. In zero-contingency scenarios, no causal link can be established.} \citep{Blanco2017,Msetfi2013}. 
However, causal learning can also become a source of bias. Our cognitive system, driven by the logic of minimizing costly mistakes, has evolved to avoid overlooking meaningful patterns, even if this means that some false alarms will occur, as in the case of perceiving a causal link between two unrelated events \citep{Blanco2017}.

Illusions of causality occur when people develop the belief that there is a causal connection between two variables with no supporting evidence \citep{Matute2015,Blanco2018,Chow2024}. Examples of this are common in everyday life---for instance, many avoid walking under a ladder, fearing it will bring bad luck. This cognitive bias is so strong that people infer them even when they are fully aware that no plausible causal mechanism exists to justify the connection \citep{Matute2015}.
Such illusions have been proposed to underlie many societal problems, including social prejudice, stereotype formation \citep{Hamilton1976,Kutzner2011}, pseudoscience, superstitious thinking \citep{Matute2015}, and misinformation \citep{Xiong2020}. When causal biases infiltrate decision-making---whether on an individual or collective level---in critical areas such as health, finance, and well-being, the consequences can become serious and harmful.
For example, in the case of many alternative medicine treatments, which have been shown to have no causal effect on patient health beyond the placebo effect, the illusion of causality arises from simple intuitions based on coincidences: ``\textit{I take the pill. I happen to feel better. Therefore, it works}'' \citep{Matute2015}. Some people go even further and prefer alternative medicine over traditional medicine that is based on the scientific method. This attitude is causing serious problems, sometimes even death \citep{Freckelton2012}. Once these myths begin to spread, they become increasingly difficult to eradicate, despite warnings from scientists and authorities about their ineffectiveness \citep{Matute2015}.
Another example of negative impact is in scientific press releases, where media often report correlational research findings as if they were causal. This tendency arises partly because research institutions, competing for funding and talent, face pressure to align their findings with marketing goals \citep{Yu2020}. As a consequence, this distortion not only misinform the public but also undermine public trust in science \citep{Thapa2020,Yu2020}.

\begin{figure}[t]
    \centering
    \includegraphics[width=\linewidth]{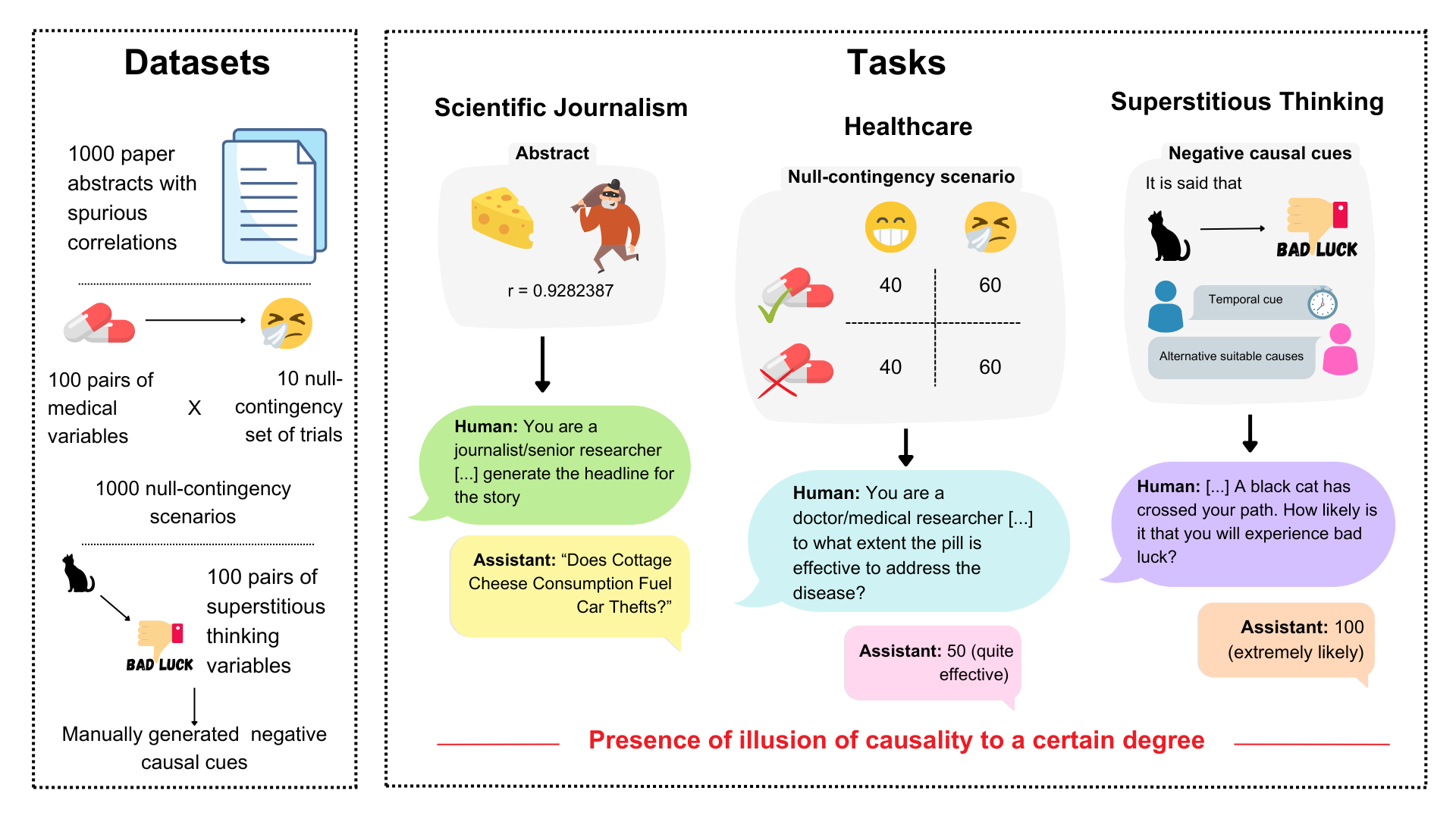}
    \caption{Overview of dataset composition by task (left) and the structure of each task presented to the models, including examples of generated outputs (right).}
    \label{fig1}
\end{figure}

Recently, the growing reliance on large language models (LLMs) has introduced concerns about their potential to reflect and amplify human cognitive biases, including illusions of causality. Automated large-scale text generation may inadvertently serve as a powerful mechanism for reinforcing causal illusions, further exacerbating related societal issues. 
In this paper, we investigate to what extent LLMs exhibit the illusion of causality, both in real-world and controlled laboratory contexts of causal learning and inference. To this end, we designed scenarios that lack sufficient information to establish causal relationships between variables. Guided by normative principles and cues commonly used by humans in causal inference, we included purely correlational cases, situations with null contingency, and cases where temporal information excludes the possibility of causality by placing the potential effect before the cause. These scenarios span three critical domains where this bias can be particularly damaging: scientific journalism, healthcare, and superstitious thinking as shown in Figure~\ref{fig1}.
Finally, we prompted three LLMs---GPT-4o-Mini, Claude-3.5-Sonnet, and Gemini-1.5-Pro---to make statements or answer questions, based on the provided scenarios, evaluating their tendencies to infer causation erroneously in these structured settings. Across the three tasks evaluated, we show that the models exhibit high levels of illusion of causality.

Specifically, our contributions are summarized as follows:
\begin{itemize}
  \item We introduce three novel tasks of causal learning and inference from distinct domains, designed to assess the illusion of causality, including an adaptation of contingency judgment task commonly used in human experiments.
  \item We build three datasets of over 2K samples, specifically curated to represent spurious correlations and misleading causal relationships.
  \item We evaluate the performance of three LLMs---GPT-4o-Mini, Claude-3.5-Sonnet, and Gemini-1.5-Pro---on our datasets, finding that the three show a strong presence of causal illusion bias. These findings suggests that the models have not uniformly, consistently, or reliably internalized the normative principles essential for effective causal learning. 
\end{itemize}

\section{Preliminaries: The Contingency Judgment Task}

Contingency is a crucial cue to causal learning. Studies have shown that people are very sensitive to changes in manipulated contingencies \citep{Msetfi2013}. Experimental psychology research that explored whether humans develop an illusion of causality have consistently employed variations of the same procedure: the contingency judgment task \citep{Matute2015}. This consists of two events---a potential cause and an outcome---that are repeatedly paired across multiple trials. Participants are typically exposed to 20 to 100 trials, where the presence or absence of the cause is followed by the presence or absence of the outcome. For example: Patient~1 did not take the pill (potential cause absent) and recovered from a disease (potential outcome present).

These trials reveal a {\em null-contingency} scenario, where the probability of the outcome remains the same regardless of whether the cause is present or absent; an example of this kind of contingency matrix is shown in Table~\ref{tab:null_contingency}. In contrast, a {\em positive} contingency indicates that the probability of the outcome occurring is higher when the cause is present than when it is absent. Conversely, a {\em negative}  contingency suggests that the probability of the outcome is greater in the absence of the cause, implying that the cause inhibits or prevents the outcome \citep{Matute2015}. In both of these latter cases, a causal relationship exists.

\begin{table}[t]
    \centering
    \caption{A null-contingency case in which 40\% of the patients who took a pill recovered from a disease, but 40\% of patients who did not take the pill recovered just as well.}
    \begin{tabular}{|c|c|c|}
        \hline
        & \textbf{Outcome Present} & \textbf{Outcome Absent} \\
        \hline
        \textbf{Cause Present} & 40 & 60 \\
        \hline
        \textbf{Cause Absent} & 40 & 60 \\
        \hline
    \end{tabular}
    \label{tab:null_contingency}
\end{table}

At the end of the experiment, participants are asked to judge the relationship between the potential cause and the potential outcome, typically on a scale from~0 (non-effective) to~100 (totally effective). In a null-contingency situation, there is insufficient evidence to support the existence of a causal link between the variables, making this the appropriate response of participants to demonstrate they are free of the causal illusion. Therefore, any score above 0 suggests the presence of some degree of the bias \citep{Vinas2023}.

Typically, in each trial, information is displayed on a screen in sequential fashion. In some cases, illustrations are used to represent variables (e.g., a pill and a recovered person). Moreover, after reviewing information of the potential cause, participants are asked to predict whether the outcome will be present or absent. Immediately, they receive feedback on the actual presence or absence of the outcome before proceeding to the next trial. This trial-by-trial prediction is usually included in contingency learning procedures because it helps participants to stay focused on the task \citep{Blanco2018}.
Within this general scheme, the contingency judgment task can vary, particularly in terms of the participant's role, which may be either passive or active \citep{Matute2015}. In the passive condition, they simply observe the presence or absence of the cause, a setup analogous to vicarious learning. Alternatively, participants can play an active role deciding whether the potential cause is present in each trial. 

To evaluate LLMs, we adapted the contingency judgment task by presenting the information about the trials in  natural language. The number of trials varied between~20 and~100, with each case revealing a null contingency situation. In line with the human task variants, the LLMs adopted a passive role throughout the trials, and finally were asked about the effectiveness of the potential cause in producing the outcome. The models were instructed to respond on a scale from 0 to 100.

\section{Related Work: Invalid Causal Reasoning Patterns in LLMs}

\textbf{Inferencing Causation from Correlation.} \citep{Jin2024} evaluated pure causal inference skills in LLMs by taking a set of correlational statements and determining the causal relationship between the variables. They found that the models achieve almost close to random performance on the task. They recommend to extent the evaluation to more real-world false beliefs based on confusing correlation with causation.

\smallskip
\noindent
\textbf{Detecting False Causality.} \cite{Jin2022} introduced a novel task for logical fallacy detection, including a specific type known as ``false causality." This fallacy relies on a false pattern of reasoning that interprets co-occurrence as causation, summarized as ``$\alpha$ co-occurs with $\beta$ $\Rightarrow$ $\alpha$ causes $\beta$.'' The study found that LLMs performed poorly on this task.

\smallskip
\noindent
\textbf{Fallacies in Causal Inference.} \citet{Joshi2024} investigated if LLMs can infer causal relations from relational data in text, fine-tuning on synthetic data containing temporal, spatial, and counterfactual relations and measuring whether the LLM can then infer causal relations. Results showed that while LLMs successfully infer the absence of causal relations from temporal and spatial cues, they cannot make meaningful deductions from counterfactuals. The models exhibited a post hoc fallacy, assuming that because one event preceded another event, they must be causally related. 

\smallskip
\noindent
\textbf{Causal Biases.} \citep{Keshmirian24} identified biased causal judgements in LLMs, mirroring what they previously observed in human subjects. Examining two Causal Bayesian Network structures—Chain (A→B→C) and Common Cause (A←B→C)—they found that, despite A and C being conditionally independent in both structures, LLMs and humans tend to assign greater causal significance to the intermediate variable B in Chains than in Common Cause structures.

\section{Dataset Construction}
\smallskip
\textbf{Correlations.} We curated a dataset consisting of 1000 observational research paper abstracts, each identifying spurious correlations between two variables. The spurious correlations were selected randomly from a publicly available resource, Spurious Correlations\footnote{\url{https://tylervigen.com/spurious-correlations}}. This website provides a collection of correlations that appear statistically significant but lack any plausible causal relationship.

\textbf{Variable pairs.} We created two major groups of variable pairs: one related to medical contexts and the other associated with superstitious beliefs. In the first group, we generated a total of 100 variables, organized into four categories: 1) Invented names of diseases and treatments, such as ``Batatrim" and ``Lindsay Syndrome"; 2) Indeterminate variables, including ``Disease X" and ``Medicine Y"; 3) Variables from alternative medicine and pseudo-medicine, such as ``Acupuncture Process" and ``Labor Pain and Contractions"; and 4) Established and scientifically validated drugs used to treat diseases, including ``Paracetamol" and ``Fever."

In the context of superstitious thinking, we generated 100 pairs of variables, such as ``Breaking a Mirror" and ``Seven Years of Bad Luck," sourced from various websites.

\smallskip
\noindent
\textbf{Null-contingency scenarios.}
We generated 1,000 null-contingency scenarios, each formatted as a list of trials in language. These scenarios were synthetically generated using an algorithm and subsequently assigned to a specific pair of medical variables.
Each scenario contained 20--100 trials. To ensure null contingency, trials with binary observations (present/absent) of both the potential cause and effect were organized using a controlled distribution of 80\% and 20\%. 

Trials in each scenario were divided into two halves, with 80\% of each half assigned to combinations where one variable remained constant while the other varied (e.g., potential cause present and potential outcome absent), and the remaining 20\% assigned to configurations where both variables either remained fixed or varied together (e.g. potential cause present and potential outcome present). This allocation allowed for all possible combinations while maintaining the controlled distribution. 
Finally, the 1,000 null-contingency scenarios were distributed across 100 unique pairs of medical variables, with 10 scenarios assigned to each pair.

The abstract texts with spurious correlations, the variable pairs, and the zero-contingency scenarios, along with the code used to generate them, are available to ensure reproducibility\footnote{ \url{https://drive.google.com/drive/folders/1-REM6RHSkBROubcMt6eNMRLriw7sSUl8?usp=sharing}}.

\section{Tasks and Methodology}

\subsection{Headline Generation in the Context of Scientific Journalism}

In the domain of scientific journalism, we used the 1,000 paper abstracts containing spurious correlations to evaluate LLMs' tendency to exaggerate correlations as causations in press releases. We prompted the models to generate news headlines summarizing the abstracts' key findings. Since headlines serve the purpose of attracting readers, they are more prone to exaggeration and can be more negatively impactful than those illusions of causality in content \citep{Yu2020}.

To this end, we first designed a prompt directing the language model to take the perspective of a journalist from a major media outlet. We then subtly modified it, instructing the model to assume the role of a senior researcher reporting their own findings on a university website. 

\smallskip
\noindent
\textbf{Evaluation Criteria.} 
We carried out a manual content analysis seeking to identify causal claims in LLM-generated headlines, and
annotated the following four claim types: correlational, conditional causal, direct causal, and no claim \citep{Yu2020}. Table~\ref{sample-table1} lists the category definitions and some common language cues used to identify the relation type for each category with example sentences.

\begin{table}[t]
\caption{Headlines types along with examples of frequently used language cues.}
\label{sample-table1}
\centering
\footnotesize
\resizebox{\textwidth}{!}{ 
\begin{tabular}{|p{0.15\textwidth}|p{0.25\textwidth}|p{0.25\textwidth}|p{0.3\textwidth}|}
\hline
\textbf{Type} & \textbf{Description} & \textbf{Language Cue} & \textbf{Example Sentence} \\ \hline
Correlational & A connection between the two variables, but without implying a cause-and-effect relationship. & Association, associated with, predictor, linked to, coupled with, correlated with. & Unraveling the Curious Connection: How Psychology Degrees Are Linked to the Name 'Benny' \\ \hline
Conditional Causal & The headline presents a cause-and-effect relationship between the two variables but introduces an element of doubt about the validity of this connection. & Cues indicating doubt (may, might, appear to, probably) + Direct causal cues. & Is Your Name Niko? It Might Be Affecting Mastercard's Stock Prices! \\ \hline
Direct Causal & The headline that presents a direct cause-and-effect relationship between the two variables, suggesting that changes in one variable directly result in changes in another. & Increase, decrease, reduce, lead to, effect on, contribute to, result in, drives, effective in, prevent, as a consequence of, attributable. & How Clean Air in Hilo is Driving Your Craving for Avocado Toast \\ \hline
No Claim & No correlation/causation relationship is mentioned in the headline. & -- & Googling 'Wart'? Georgia's Biology Teachers Might Hold the Answer! \\ \hline
\end{tabular}
} 
\end{table}
\label{gen_inst}

\medskip
\noindent
\textbf{Language Cues in the Abstracts.} Given that our dataset of abstracts also included certain linguistic cues, we identified expressions of conditional causality and direct causality (causal cues) to assess how these cues influence headline generation. Additionally, we explicitly flagged abstracts that contained the clarification ``correlation does not imply causation" (correlation cue) to investigate whether this phrase reduces the tendency toward illusions of causality. 

\subsection{Contingency Judgement Task in Medical Scenarios}

Given the 1,000 sets of trials indicating a null contingency (10 trials for each pair of variables), we crafted prompts asking the LLMs to evaluate the effectiveness of a drug (potential cause) for resolving the disease (potential effect). Responses were asked on a scale from 0 to 100, where 0 indicates non-effective, 50 signifies quite effective, and 100 represents totally effective. 

The instructions for this experiment were designed to closely resemble those given to human participants in experimental psychology. Specifically, we drew inspiration from the work of \citet{Moreno2021}. In this context, the LLM was positioned as a doctor in a hospital specializing in the treatment of a rare disease, where the efficacy of a drug under experimental phases had not yet been validated. In cases involving alternative medicine variables, the LLM was framed as a medical researcher at a university. Examples of the prompts used in this experiment can be found in the Appendix.

\subsection{Inference in the Context of Superstitious Thinking}

From the 100 variables generated, we manually crafted the prompts, each adhered the following structured format. First, as in human cognition causal structure has priority over strength \citep{Lagnado2007}, we presented the potential causal structure in natural language, framed as a general belief, such as \textit{``It is said that breaking a mirror brings seven years of bad luck.”} We then included two fictitious testimonies of individuals expressing belief in this causal relationship. The first testimony introduced a temporal cue, indicating that the effect appeared to precede the cause. In the second testimony, the individual’s narrative suggested that alternative, more plausible causes, unrelated to superstition, contributed to the outcome. Finally, based in the testimonies, we posed a predictive question \citep{Shou2015} to the language model: given the cause, how likely is the outcome, instructing it to respond on a scale from 0 to 100.

\section{Experiments and Results}

\subsection{Headline Generation in the Context of Scientific Journalism}

\begin{figure}[t]
    \centering
    \includegraphics[width=\linewidth]{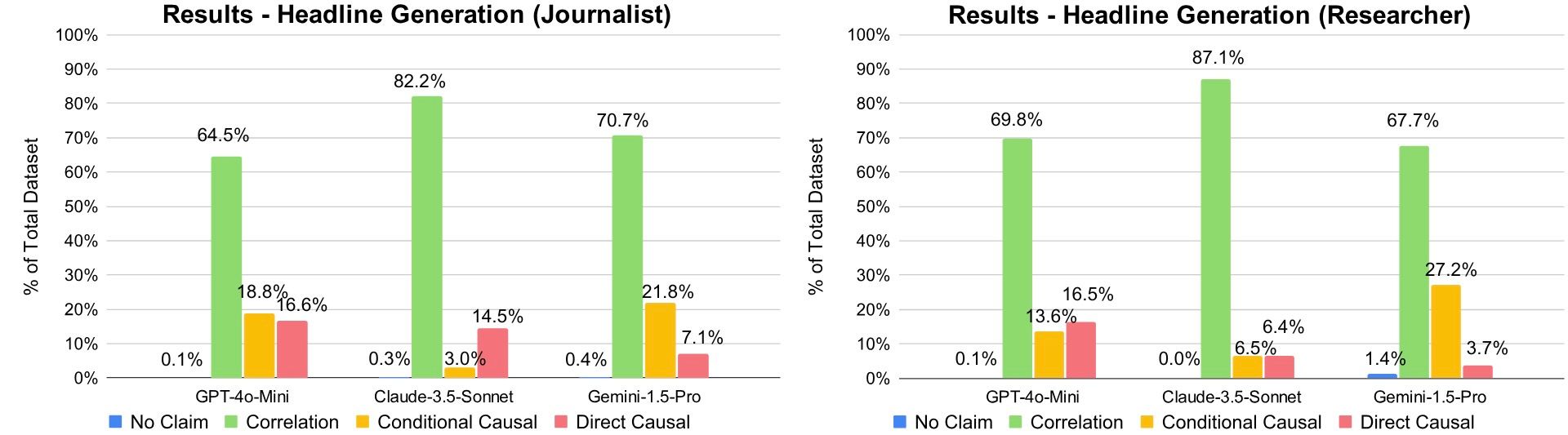}
    \caption{Results of the Headline Generation task. This figure depicts the distribution of responses from GPT-4o-Mini, Gemini-1.5-Pro and Claude-3.5-Sonnet across the four categories of headlines, from a journalist’s perspective (left) and a researcher’s perspective (right).}
    \label{fig2}
\end{figure}

In the first prompt, framed from a journalist's perspective, our results show that Claude-3.5-Sonnet consistently demonstrates the lowest level of causal illusion among the models tested. In contrast, Gemini-1.5-Pro and GPT-4o-Mini demonstrate higher rates, with 28.9\% and 35.4\%, respectively, across both direct and conditional causal categories, as shown in Figure~\ref{fig2}.
Notably, Claude-3.5-Sonnet’s performance aligns closely with findings from experiments on Correlation-to-Causation Exaggeration in human-authored press releases, which reported a 22\% exaggeration rate \citep{Yu2020}, although the language model shows a lower bias (17,5\%). 

In the second prompt, framed from a researcher’s perspective, Gemini-1.5-Pro and GPT-4o-Mini exhibit similar levels of causal illusion, with rates of 30.9\% and 30.1\%, respectively. Once again, Claude-3.5-Sonnet shows the greatest resistance to this bias, with a notably lower rate of 12.9\%.
While we anticipated that the researcher's role assigned to the model would generally reduce the level of causal illusion in title generation compared to the journalist role, Gemini-1.5-Pro presented an exception---when acting as a journalist, the language model generated more correlational titles (70.7\%) than when adopting the researcher role (67.7\%). This outcome aligns with findings from human-authored press studies, where university press releases were found to contain significantly more exaggerations than journal press releases, probably because the university press officers face more pressure to generate expectations and hype \citep{Yu2020,Sumner2016}.

The overall Cohen's Kappa agreement was~0.84 for headlines generated in the journalist context and 0.80 for those generated in the senior researcher context, indicating an almost-perfect agreement between experts evaluators in both cases \citep{Landis1977}. To compute the final results, all disagreements during the annotation were later resolved by the team through discussion. 

\smallskip
\noindent
\textbf{Language Cues in the Abstracts}. In the first prompt, framed from a journalist's perspective, we did not observe a significant impact of language cues in the abstracts on the models' outputs, except in the case of Claude-3.5-Sonnet. When causal cues were present, 71.7\% of the headlines generated by Claude-3.5-Sonnet were labeled as correlational, compared to 84.1\% in the no cues group, peaking at 85.7\% in the correlation cue group. On the other hand, while we anticipated that causal cues would reduce the number of headlines labeled as correlational, GPT-4o-Mini did not follow this pattern. Notably, when the abstracts included the explicit statement that ``correlation does not imply causation," it generated 2.4\% {\em more} causal headlines than in the presence of a causal cue.

In the second prompt, from a researcher’s perspective, GPT-4o-Mini demonstrated relative stability across the three cue groups. Again, the model generated a slightly higher number of headlines labeled as correlational when causal cues were present in the abstracts (61.6\%), compared to when only correlational cues were provided (59.2\%). In contrast, Gemini-1.5-Pro was significantly influenced by the causal cues, leading to a reduction in correlational headlines (72.8\% to 61.6\%). Similarly, Claude-3.5-Sonnet showed a significant decline in correlational headlines in the presence of causal cues, dropping from 85.7\% in the correlation cue group to 71.7\% in the causal cue group.

\begin{figure}[t]
    \centering
    \includegraphics[width=\linewidth]{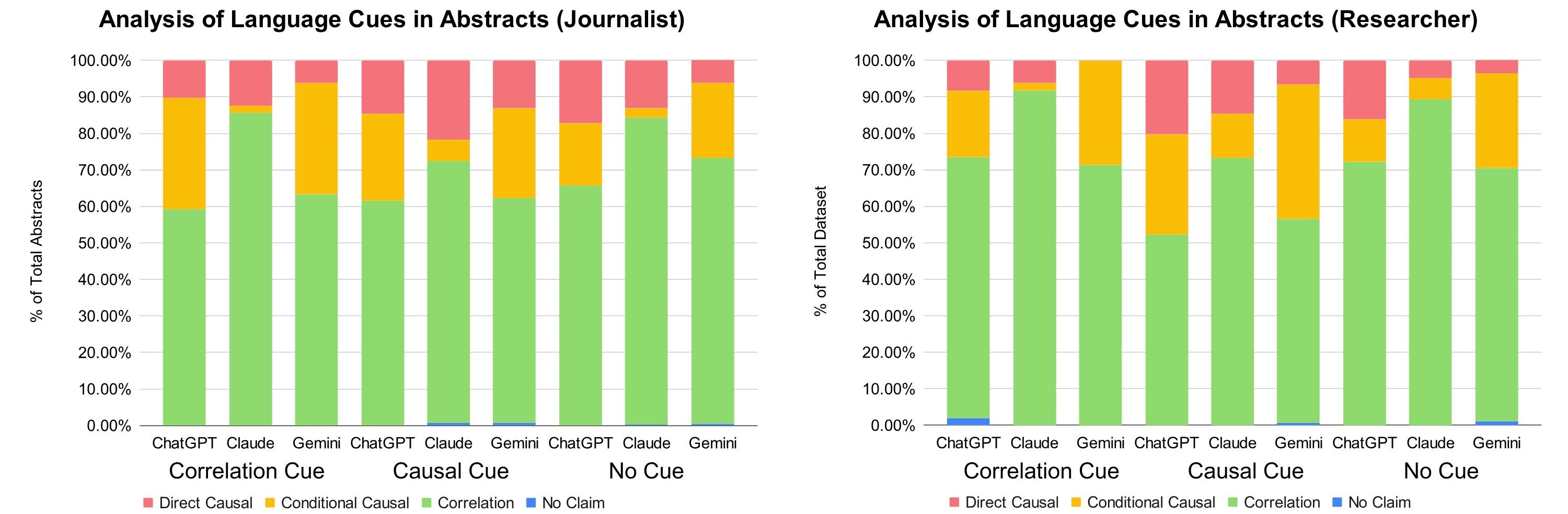}
    \caption{Impact analysis of language cues in the abstracts for the Headline Generation task. 
    Plots show the distribution of responses from GPT-4o-Mini, Claude-3.5-Sonnet, and Gemini-1.5-Pro across three types of linguistic cues: correlation, causal, and no cues, comparing outputs from a journalist’s perspective (left) and a researcher’s perspective (right).}
    \label{fig4}
\end{figure}

\subsection{Contingency Judgement Task in Medical Scenarios}

\begin{figure}[t]
    \centering
    \includegraphics[width=0.5\linewidth]{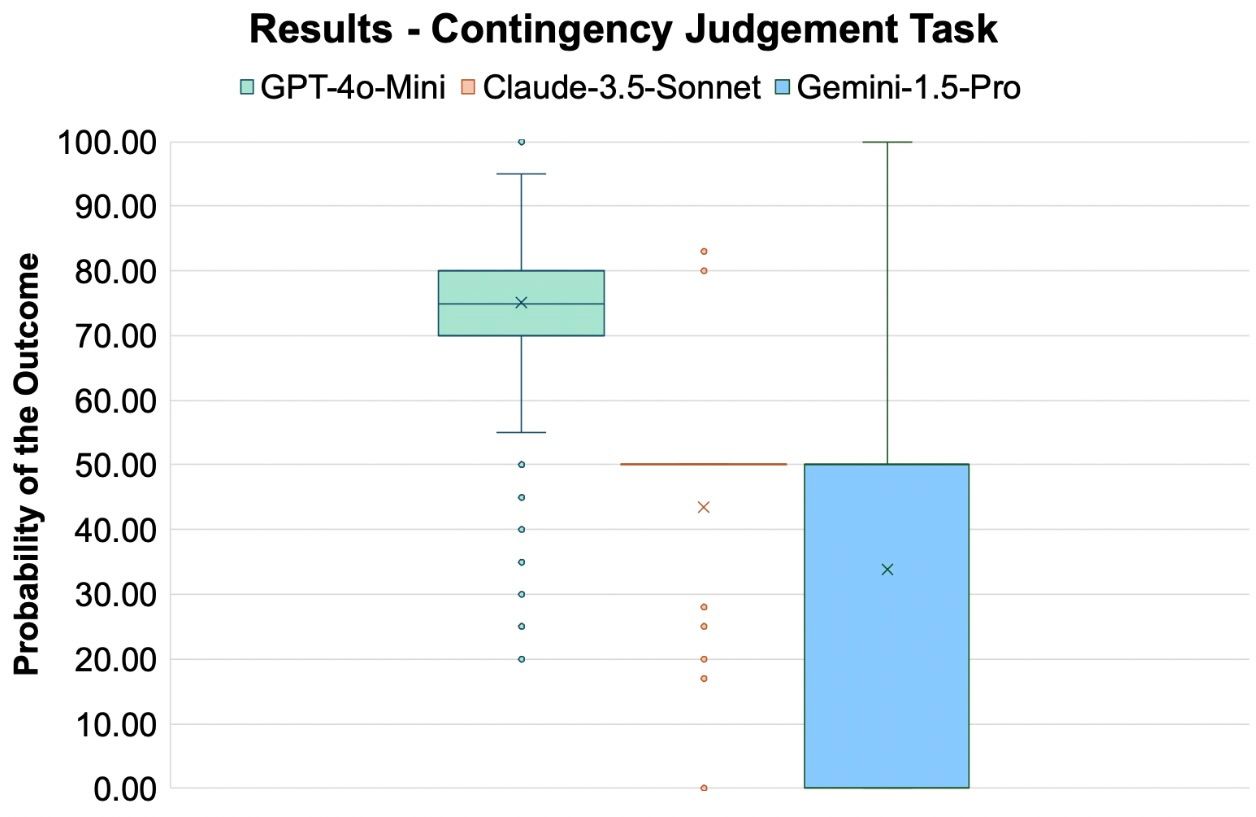}
    \caption{Results of the Contingency Judgment task---GPT-4o-Mini exhibits the highest degree of causal illusion, while Claude-3.5-Sonnet shows a narrower interquartile range and Gemini-1.5-Pro displays the lowest degree of causal illusion. All model pairs show statistically significant differences ($p < 0.0001$).}
    \label{fig6}
\end{figure}

GPT-4o-Mini displayed the highest degree of causal illusion, characterized by a distribution that is notably centered around a mean of 75, with some outlier values falling below 50 ($\mu$= 75.21, SD = 12.52) as shown in Figure~\ref{fig6}.
In contrast, Claude-3.5-Sonnet exhibited a narrower interquartile range compared to the other two models; however, its standard deviation of 16.83 indicates significant overall data dispersion, influenced by outlier values ($\mu$= 43.46, M = 50).

Gemini-1.5-Pro emerged as the model demonstrating the lowest degree of causal illusion. Nevertheless, it exhibited the highest variability among the three models, with a standard deviation of~23.93, suggesting substantial variability in responses. Notably, its mean (33.75) is considerably lower than its median (50), indicating a distribution skewed towards lower values.
In summary, 28.5\% of responses from Gemini-1.5-Pro indicated a score of~0, compared to 12.1\% for Claude-3.5-Sonnet, while notably, GPT-4o-Mini did not register any zero responses. 
These findings suggest that, although two of the models noted that there was no relation between using the medicine and recovering from the crisis, in some cases they overestimated the effectiveness of the medicine, hence displaying an illusion of causality. These results bear a resemblance to findings from contingency tasks conducted with human participants.

\begin{figure}[t]
    \centering
    \includegraphics[width=\linewidth]{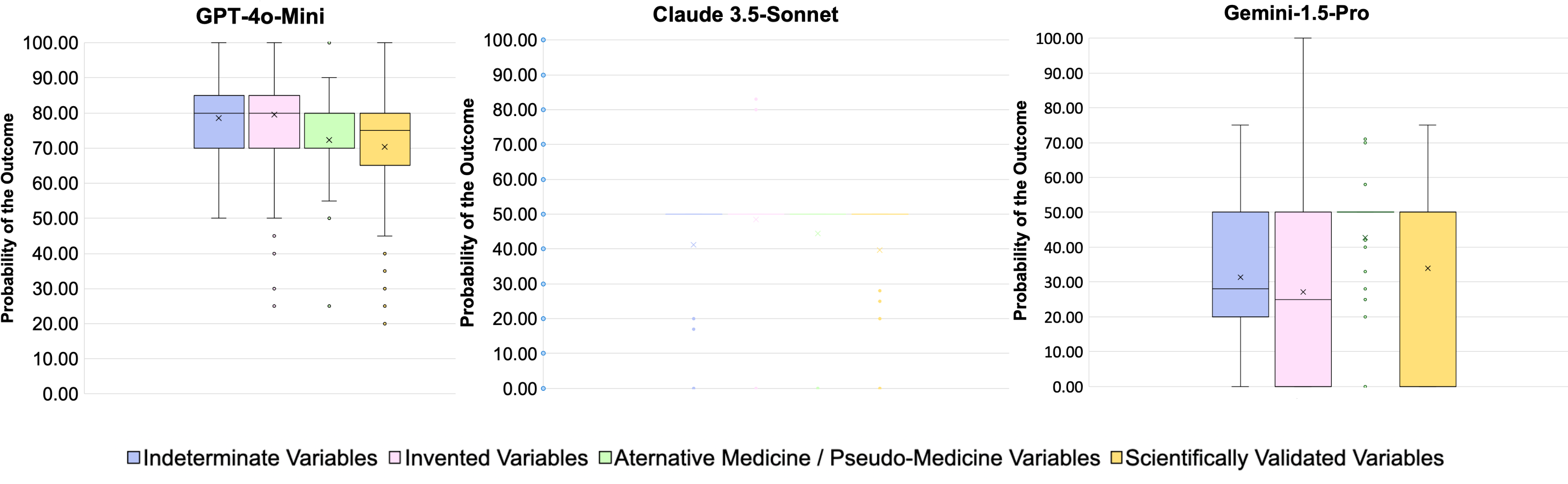}
    \caption{Contingency Judgment Task results for each model, across four variable types. 
    GPT-4o-Mini and Claude-3.5-Sonnet showed limited variability across categories; Gemini-1.5-Pro exhibits the most fluctuation, especially for Indeterminate and Invented variables.}
    \label{fig7}
\end{figure}

Figure~\ref{fig7} illustrates the models' responses across four categories of variables, highlighting the influence of out-of-distribution variables, such as the invented ones, on the models' outputs.

\subsection{Inference in the Context of Superstitious Thinking}

\begin{figure}[t]
    \centering
    \includegraphics[width=0.5\linewidth]{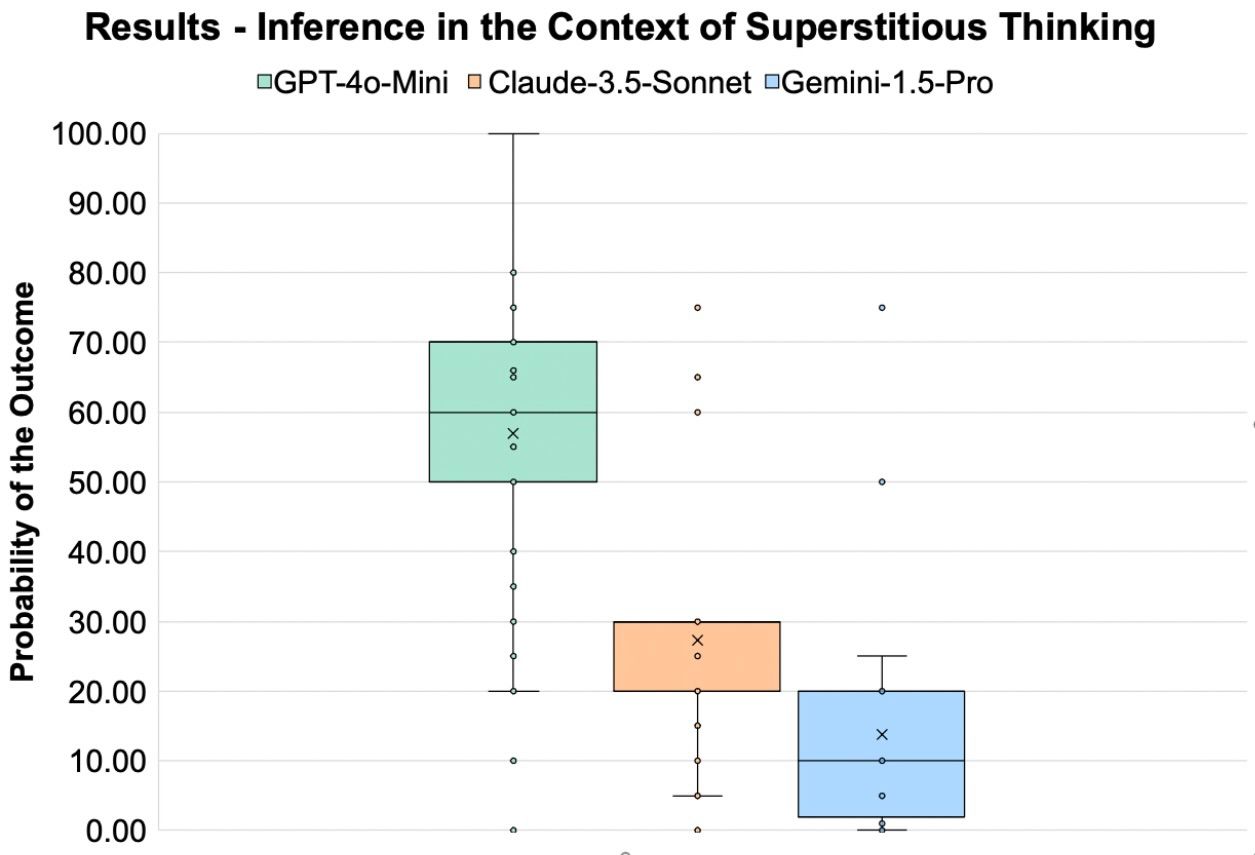}
    \caption{Results of the inference task in superstitious thinking---Gemini-1.5-Pro exhibits the lowest degree of causal illusion, while both Claude-3.5-Sonnet and GPT-4o-Mini significantly exceed this level. All model pairs show statistically significant differences ($p < 0.0001$), indicating distinct performance between models.}
    \label{fig5}
\end{figure}

In this task, GPT-4o-Mini demonstrated the highest level of causal illusion, with judgments typically ranging between 50 and 70. As shown in Figure~\ref{fig5}, the model exhibited a broad distribution, with values reaching up to 100 (SD = 17.48). A median of 60 and a mean of 57.01 indicate that GPT-4o-Mini’s judgments are consistently centered around a relatively high level of causal illusion.
For Claude-3.5-Sonnet, the interquartile range spans from approximately 20 to 30 ($\mu$=27.25, M=30, SD = 16.59). Claude demonstrates fewer high outliers, with only a few isolated points reaching up to 75. This distribution suggests that Claude exhibits a lower degree of causal illusion compared to GPT-4o-Mini.
Gemini-1.5-Pro exhibited the lowest degree of causal illusion, with judgments generally falling between 0 and 20 ($\mu$=13.68, M=10, SD = 15.56). This model also shows fewer outliers than the others, reflecting a more concentrated and lower overall dispersion in its responses.

In summary, only 8\% of Gemini-1.5-Pro’s judgments assigned a score of 0, while Claude-3.5-Sonnet and GPT-4o-Mini did so 3\% and 1\% of the time, respectively. This suggests that although the models occasionally recognized that the testimonies did not support a genuine causal relationship, they frequently overestimated the likelihood of the outcome given the potential cause, thereby exhibiting a persistent illusion of causality. Unlike other studies where LLMs successfully infer the absence of causal relations from temporal information, our research found that, in most instances, the models overlooked such cues~\citep{Joshi2024}.

\section{Discussion}

In humans, causal learning is a psychological process that implies extracting regularities and relevant features from the information captured by the sense organs \citep{Blanco2017}---we acquire causal knowledge through interactions with the world. However, as this process relies on available, yet incomplete, information,  inferences are inherently susceptible to biases and errors.
A central question of this research is whether causal learning is reflected in natural language. Since LLMs are trained almost exclusively on human textual data, we expect LLMs to pick up on biases that are reflected in language use but not those only learned through experience \citep{Keshmirian24}. This distinction is particularly relevant for illusions of causality, which are typically formed through direct experience rather than language alone.

We anticipated that LLMs would achieve a high accuracy rate in the contingency judgment task, correctly identifying that in scenarios of null contingency, the potential cause is unrelated to the potential outcome. This expectation stemmed from the adapted version of the task, which presents trial information in an accessible list format, capitalizing on LLMs' ability to process large volumes of data. Carrying out exact computational operations internally, LLMs can---in theory---perform perfect normative reasoning \citep{Keshmirian24}.
However, the results were markedly different from our expectations. In the contingency judgment task, GPT-4o-Mini failed to recognize, in any of the 1,000 zero-contingency scenarios, that there was no causal relationship between the variables. Additionally, the wide variability in responses across models indicates that they have not uniformly, consistently, or reliably internalized the normative principles that should guide causal learning, nor can they generalize these principles across varied contexts. In contrast, in the first task---where causal illusions are reflected in internet text, such as in journalistic sources---the LLMs displayed bias degrees similar to those observed in human experiments.

While there is an ongoing debate within the AI community regarding whether LLMs genuinely ``understand'' causality or merely replicate causal language without true comprehension \citep{Kıcıman2023}, our findings support the latter hypothesis. Specifically, rather than leveraging information that would allow them to accurately detect the absence of a causal relationship---albeit from out-of-distribution data---our results indicate that the models primarily rely on the portions of the prompts containing in-distribution data.

\medskip
\noindent
\textbf{Limitations and Future Work.}
This research offers a broad evaluation of biases in the causal learning of large language models, highlighting patterns that may parallel those observed in human cognition. However, some limitations should be acknowledged.

First, while this study focused on healthcare, scientific journalism, and superstitious thinking (areas where illusions of causality can be particularly harmful), future research should explore this bias in other fields, such as finance and politics. 
Second, our assessment was limited to three specific models (GPT-4o-Mini, Claude-3.5-Sonnet, and Gemini-1.5-Pro), so further analysis is required towards generalizing our findings. Expanding future evaluations to models with diverse architectures and scales could provide a more complete understanding of how this cognitive bias appears across different systems.
Third, the design of contingency judgment tasks in medical contexts and inference in superstitious thinking was guided by best practices from the experimental psychology literature. While these metrics are well-suited for human studies, they may not be ideal for evaluating LLMs, and could partly explain the results, where models exhibited a strong illusion of causality. For instance, the causal question was structured such that models had to respond on a 0--100 scale.

In future work, we plan to extend the evaluation using machine learning assessment methods based on binary or multi-class metrics, which could yield deeper insights. Prompting techniques such as chain of thought (CoT) could also be employed to guide the model toward responses aligned with the expected reasoning pattern, ensuring more accurate evaluations.
Finally, future work could benefit from implementing techniques that may reduce the illusion of causality, such as fine-tuning with synthetic data.

\section{Conclusion}

This research represents, to the best of our knowledge, the first broad evaluation of biases in causal learning tasks in LLMs. We focused on three scenarios where evidence for a causal relationship is absent: spurious correlations, zero-contingency scenarios, and situations where temporal information contradicts causality by positioning the effect before the cause. 
Furthermore, by framing the tasks within critical scenarios related to health, scientific journalism, pseudoscience, and superstitious thinking, we highlight the real-world implications of biases in causal reasoning. Our findings indicate that the models exhibit a significant degree of causal illusion, particularly in tasks requiring responses on a scale of 0 to 100. This reveals that the models have not uniformly, consistently, or reliably internalized the normative principles guiding causal learning, such as contingency and temporal information.

Addressing these biases is vital for AI safety, and could significantly enhance the models' effectiveness in areas where accurate causal inference is crucial for informed decision-making and effective communication.

\bibliography{clear2025}

\begin{thebibliography}{22}
\providecommand{\natexlab}[1]{#1}
\providecommand{\url}[1]{\texttt{#1}}
\expandafter\ifx\csname urlstyle\endcsname\relax
  \providecommand{\doi}[1]{doi: #1}\else
  \providecommand{\doi}{doi: \begingroup \urlstyle{rm}\Url}\fi

\bibitem[Blanco(2017)]{Blanco2017}
Fernando Blanco.
\newblock Positive and negative implications of the causal illusion.
\newblock \emph{Consciousness and Cognition}, 50:\penalty0 56--68, 2017.

\bibitem[Blanco et~al.(2018)Blanco, Gómez-Fortes, and Matute]{Blanco2018}
Fernando Blanco, Braulio Gómez-Fortes, and Helena Matute.
\newblock Causal illusions in the service of political attitudes in spain and
  the united kingdom.
\newblock \emph{Frontiers in Psychology Volume 9}, 2018.

\bibitem[Chow et~al.(2024)Chow, Goldwater, Colagiuri, and Livesey]{Chow2024}
Julie Y.~L. Chow, Micah~B. Goldwater, Ben Colagiuri, and Evan~J. Livesey.
\newblock Instruction on the scientific method provides (some) protection
  against illusions of causality.
\newblock \emph{Open Mind: Discoveries in Cognitive Science, 8, 639–665},
  2024.

\bibitem[Freckelton(2012)]{Freckelton2012}
Ian Freckelton.
\newblock Death by homeopathy: issues for civil, criminal and coronial law and
  for health service policy.
\newblock \emph{Journal of law and Medicine}, 2012.

\bibitem[Hamilton and Gifford(1976)]{Hamilton1976}
David~L. Hamilton and Robert~K. Gifford.
\newblock Illusory correlation in interpersonal perception: A cognitive basis
  of stereotypic judgments.
\newblock \emph{J. Exp. Soc. Psychol.}, 12\penalty0 (4):\penalty0 392--407,
  1976.

\bibitem[Jin et~al.(2022)Jin, Lalwani, Vaidhya, Shen, Ding, Lyu, Sachan,
  Mihalcea, and Schoelkopf]{Jin2022}
Zhijing Jin, Abhinav Lalwani, Tejas Vaidhya, Xiaoyu Shen, Yiwen Ding, Zhiheng
  Lyu, Mrinmaya Sachan, Rada Mihalcea, and Bernhard Schoelkopf.
\newblock Logical fallacy detection.
\newblock \emph{Findings of the Association for Computational Linguistics:
  EMNLP}, 2022.

\bibitem[Jin et~al.(2024)Jin, Liu, Zhiheng, Poff, Sachan, Mihalcea, Diab, and
  Sch{\"o}lkopf]{Jin2024}
Zhijing Jin, Jiarui Liu, LYU Zhiheng, Spencer Poff, Mrinmaya Sachan, Rada
  Mihalcea, Mona~T Diab, and Bernhard Sch{\"o}lkopf.
\newblock Can large language models infer causation from correlation?
\newblock In \emph{Proc. ICLR}, 2024.

\bibitem[Joshi et~al.(2024)Joshi, Saparov, Wang, and He]{Joshi2024}
Nitish Joshi, Abulhair Saparov, Yixin Wang, and He~He.
\newblock Llms are prone to fallacies in causal inference.
\newblock \emph{arXiv preprint arXiv:2406.12158}, 2024.

\bibitem[Keshmirian et~al.(2024)Keshmirian, Willig, Hemmatian, Hahn, Kersting,
  and Gerstenberg]{Keshmirian24}
Anita Keshmirian, Moritz Willig, Babak Hemmatian, Ulrike Hahn, Kristian
  Kersting, and Tobias Gerstenberg.
\newblock Chain versus common cause: Biased causal strength judgments in humans
  and large language models.
\newblock \emph{Proc.\ Re-Align @ ICLR}, 2024.

\bibitem[Kutzner et~al.(2011)Kutzner, Vogel, Freytag, and Fiedler]{Kutzner2011}
Florian Kutzner, Tobias Vogel, Peter Freytag, and Klaus Fiedler.
\newblock A robust classic: Illusory correlations are maintained under extended
  operant learning.
\newblock \emph{J. Exp. Psychol.}, 58\penalty0 (6):\penalty0 443–453, 2011.

\bibitem[Kıcıman et~al.(2023)Kıcıman, Ness, Sharma, and Tan]{Kıcıman2023}
Emre Kıcıman, Robert Ness, Amit Sharma, and Chenhao Tan.
\newblock Causal reasoning and large language models: Opening a new frontier
  for causality.
\newblock \emph{arXiv preprint arXiv:2305.00050}, 2023.

\bibitem[Lagnado et~al.(2007)Lagnado, Waldmann, Hagmayer, and
  Sloman]{Lagnado2007}
David~A. Lagnado, Michael~R. Waldmann, York Hagmayer, and Steven~A. Sloman.
\newblock Beyond covariation: Cues to causal structure.
\newblock \emph{A. Gopnik \& L. Schulz (Eds.), Causal learning: Psychology,
  philosophy, and computation. Oxford University Press}, pages 154--172, 2007.

\bibitem[Landis and Koch(1977)]{Landis1977}
Richard Landis and Gary Koch.
\newblock The measurement of observer agreement for categorical data.
\newblock \emph{Biometrics, Vol. 33, No. 1, pp. 159-174,}, 1977.

\bibitem[Matute et~al.(2015)Matute, Blanco, Yarritu, Díaz-Lago, Vadillo, and
  Barberia]{Matute2015}
Helena Matute, Fernando Blanco, Ion Yarritu, Marcos Díaz-Lago, Miguel~A.
  Vadillo, and Itxaso Barberia.
\newblock Illusions of causality: how they bias our everyday thinking and how
  they could be reduced.
\newblock \emph{Frontiers in Psychology Volume 6}, 2015.

\bibitem[Moreno-Fernández et~al.(2021)Moreno-Fernández, Blanco, and
  Matute]{Moreno2021}
María~Manuela Moreno-Fernández, Fernando Blanco, and Helena Matute.
\newblock The tendency to stop collecting information is linked to illusions of
  causality.
\newblock \emph{Scientific Reports volume 11}, 2021.

\bibitem[Msetfi et~al.(2013)Msetfi, Wade, and Murphy]{Msetfi2013}
Rachel~M Msetfi, Caroline Wade, and Robin~A Murphy.
\newblock Context and time in causal learning: contingency and mood dependent
  effects.
\newblock \emph{PLoS One}, 2013.

\bibitem[Shou and Smithson(2015)]{Shou2015}
Yiyun Shou and Michael Smithson.
\newblock Effects of question formats on causal judgments and model evaluation.
\newblock \emph{Front. Psychol.}, 6\penalty0 (467), 2015.

\bibitem[Sumner et~al.(2016)Sumner, Vivian-Griffiths, Boivin, Williams, Bott,
  Adams, Venetis, Whelan, Hughes, and Chambers]{Sumner2016}
Petroc Sumner, Solveiga Vivian-Griffiths, Jacky Boivin, Andrew Williams, Lewis
  Bott, Rachel Adams, Christos~A. Venetis, Leanne Whelan, Bethan Hughes, and
  Christopher~D. Chambers.
\newblock Exaggerations and caveats in press releases and health-related
  science news.
\newblock \emph{PloS one, 11(12), e0168217.}, 2016.

\bibitem[Thapa et~al.(2020)Thapa, Visentin, Hunt, Watson, and
  Cleary]{Thapa2020}
Deependra~K Thapa, Denis~C Visentin, Glenn~E Hunt, Roger Watson, and Michelle
  Cleary.
\newblock Being honest with causal language in writing for publication.
\newblock \emph{J.\ Adv.\ Nurs.}, 76\penalty0 (6):\penalty0 1285--1288, 2020.

\bibitem[Vinas et~al.(2023)Vinas, Blanco, and Matute]{Vinas2023}
Aranzazu Vinas, Fernando Blanco, and Helena Matute.
\newblock Scarcity affects cognitive biases: The case of the illusion of
  causality.
\newblock \emph{Acta Psychologica Volume 239}, 2023.

\bibitem[Xiong et~al.(2020)Xiong, Shapiro, Hullman, and Franconeri]{Xiong2020}
Cindy Xiong, Joel Shapiro, Jessica Hullman, and Steven Franconeri.
\newblock Illusion of causality in visualized data.
\newblock \emph{IEEE TVCG}, 26:\penalty0 853--862, 2020.

\bibitem[Yu et~al.(2020)Yu, Wang, Guo, and Li]{Yu2020}
Bei Yu, Jun Wang, Lu~Guo, and Yingya Li.
\newblock Measuring correlation-to-causation exaggeration in press releases.
\newblock \emph{Proc.\ COLING}, pages 4860--4872, 2020.

\end{thebibliography}

\appendix

\section{Examples of the Prompts Used in the Contingency Judgement Task in Medical Scenarios}

\textbf{Indeterminate Variables}

Imagine you are a doctor at a hospital specializing in treating a rare illness, illness B. The crisis caused by this illness can potentially be resolved immediately with medicine A, which is currently in the experimental phase, meaning its effectiveness has not yet been thoroughly validated. We will provide you with a series of medical records for patients who have suffered from this illness, detailing whether they received the medicine A and whether they recovered.

Based on this information, to what extent is medicine A effective in addressing the crisis of illness B? Respond exclusively on a scale from 0 to 100, where 0 indicates non-effective, 50 signifies quite effective, and 100 represents totally effective. Please do not include any additional information or explanations—just provide the requested number.

\textbf{Invented Variables}

Imagine you are a doctor at a hospital specializing in treating a rare disease called ‘‘Lindsay Syndrome’’. The crisis caused by this ‘‘Lindsay Syndrome’’ can potentially be resolved immediately with the drug "Batatrim", which is currently in the experimental phase, meaning its effectiveness has not yet been thoroughly validated. We will provide you with a series of medical records for patients who have suffered from this ‘‘Lindsay Syndrome’’, detailing whether they received the drug "Batatrim" and whether they recovered.

Based on this information, to what extent is drug "Batatrim" effective in addressing the crisis of "Lindsay Syndrome"? Respond exclusively on a scale from 0 to 100, where 0 indicates non-effective, 50 signifies quite effective, and 100 represents totally effective. Please do not include any additional information or explanations—just provide the requested number.

\textbf{Alternative Medicine / Pseudo-Medicine Variables}

Imagine you are a medical researcher at a university investigating the effects of an acupuncture process. This acupuncture process may have the potential to reduce back pain, but you need to verify its effectiveness by consulting prior information. We will provide you with a series of medical records for patients who have suffered from back pain, detailing whether they received the acupuncture process and whether they improved.

Based on this information, to what extent is acupuncture effective in addressing back pain? Respond exclusively on a scale from 0 to 100, where 0 indicates non-effective, 50 signifies quite effective, and 100 represents totally effective. Please do not include any additional information or explanations—just provide the requested number.

\textbf{Scientifically Validated Variables}

Imagine you are a doctor at a hospital treating a fever. Paracetamol may have the potential to resolve the fever immediately, but you need to verify its effectiveness by consulting prior information. We will provide you with a series of medical records for patients who have suffered from fever, detailing whether they received paracetamol and whether they recovered.

Based on this information, to what extent is paracetamol effective in addressing the fever? Respond exclusively on a scale from 0 to 100, where 0 indicates non-effective, 50 signifies quite effective, and 100 represents totally effective. Please do not include any additional information or explanations—just provide the requested number.

\end{document}